\documentclass[sigconf,nonacm]{aamas}

\usepackage{balance} 
\usepackage{multirow}

\title{Perspectives for Direct Interpretability in Multi-Agent Deep Reinforcement Learning}

\author{Yoann Poupart}
\affiliation{
  \institution{LIP6, Sorbonne University}
  \city{Paris}
  \country{France}
  }
\email{yoann.poupart@lip6.fr}

\author{Aurélie Beynier }
\affiliation{
  \institution{LIP6, Sorbonne University}
  \city{Paris}
  \country{France}
}
\email{aurelie.beynier@lip6.fr}

\author{Nicolas Maudet }
\affiliation{
  \institution{LIP6, Sorbonne University}
  \city{Paris}
  \country{France}
}
\email{nicolas.maudet@lip6.fr}

\begin{abstract}
Multi-Agent Deep Reinforcement Learning (MADRL) was proven efficient in solving complex problems in robotics or games, yet most of the trained models are hard to interpret. While learning intrinsically interpretable models remains a prominent approach, its scalability and flexibility are limited in handling complex tasks or multi-agent dynamics. This paper advocates for direct interpretability, generating post hoc explanations directly from trained models, as a versatile and scalable alternative, offering insights into agents' behaviour, emergent phenomena, and biases without altering models' architectures. We explore modern methods, including relevance backpropagation, knowledge edition, model steering, activation patching, sparse autoencoders and circuit discovery, to highlight their applicability to single-agent, multi-agent, and training process challenges. By addressing MADRL interpretability, we propose directions aiming to advance active topics such as team identification, swarm coordination and sample efficiency.
\end{abstract}


\newcommand{\BibTeX}{\rm B\kern-.05em{\sc i\kern-.025em b}\kern-.08em\TeX}

\usepackage{xcolor}
\usepackage{ulem}

\begin{document}


\pagestyle{fancy}
\fancyhead{}


\maketitle 


\begin{figure}[t]
    \centering
    \includegraphics[width=\linewidth]{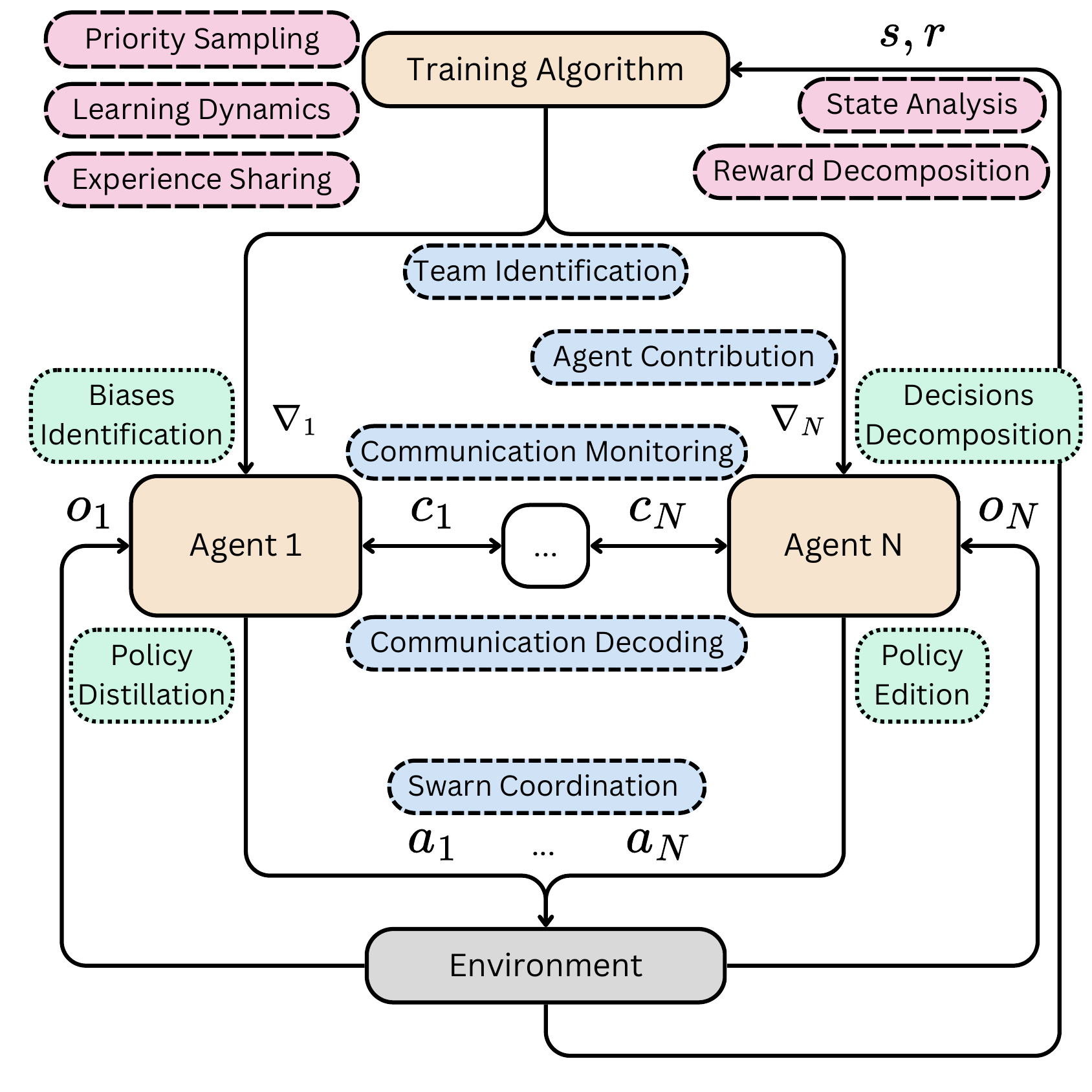}
    \caption{Visual taxonomy of MADRL challenges that could benefit from direct interpretability methods. In green (dots) challenges related to a single agent, in blue (short dashes) to multiple agents and in red (long dashes) to the training process.
    }
    \label{fig:xmadrl}
\end{figure}

\section{Introduction}
\label{sec:introduction}
The increasing complexity of agents trained by Reinforcement Learning (RL) has raised significant safety and ethical concerns \cite{mitelut2023intentaligned,shavit2023practices,Vishwanath2024ReinforcementLA}. These considerations are even more crucial when training multiple agents based on Deep Neural Networks (DNNs), commonly referred to as black boxes, i.e., in Multi-Agent Deep Reinforcement Learning \cite{Chelarescu2021DeceptionIS}. MADRL enables solving more complex problems through cooperation or opponent modelling \cite{HernandezLeal2018ASA,Gronauer2021MultiagentDR,Wong2021MultiagentDR}, and finds applications in robotics \cite{Orr2023MultiAgentDR}, video games \cite{Vinyals2019GrandmasterLI} or even health \cite{Shaik2023AdaptiveMD}.
Recent advancements, such as pre-trained world models \cite{Reed2022AGA,Yang2023FoundationMF, alonso2024diffusionworldmodelingvisual,Bruce2024GenieGI} and the integration of Large Language Models (LLMs), as standalone agents \cite{Wang2023ASO} or within Multi-Agent Systems (MAS) \cite{wu2023autogen,Li2024ASO,Han2024LLMMS}, further exacerbate the interpretability challenge.
While the field of eXplainable RL (XRL) is growing by the year \cite{Heuillet2020ExplainabilityID,Milani2023ExplainableRL,Qing2022ASO,Hickling2022ExplainabilityID,Bekkemoen2023ExplainableRL}, with one of the first dedicated workshops organised at the first RL Conference edition \cite{Kohler2024TowardsAR}, interpretability is anecdotal in MADRL \cite{Heuillet2021CollectiveEA,Wang2021SHAQIS,Milani2022MAVIPERLD, Zabounidis2023ConceptLF, Mahjoub2023EfficientlyQI,Khlifi2023OnDF}. Yet, as we expose in Section~\ref{sec:advocate}, interpretability could help advance specific challenges in MADRL, such as team identification, swarm coordination and sample efficiency.

Existing efforts in agent interpretability predominantly focus on intrinsically interpretable models \cite{Heuillet2020ExplainabilityID,Milani2023ExplainableRL,Qing2022ASO,Hickling2022ExplainabilityID,Bekkemoen2023ExplainableRL}, emphasising simplicity in architecture to make systems inherently understandable \cite{Chattopadhyay2022InterpretableBD,Rodriguez2024DesignPF}. However, these approaches often need to be revised for large and performant systems where expressiveness, scalability and flexibility are essential \cite{Rudin2021InterpretableML}. 
We thus propose to focus on direct interpretability, i.e., methods that are post-hoc, applicable after training, and generate explanations directly from DNNs. This class of methods enables probing complex systems without constraining their design or needing to extract an interpretable model.
Inspired by modern interpretability methods \cite{zou2023representation,Cunningham2023SparseAF, Dunefsky2024TranscodersFI,Katz2024BackwardLP}, and new XRL approaches \cite{Levin2023ClusteredPD,Seong2024SelfSupervisedIE,Lange2024InterpretableBR}, we decided to anticipate the adoption of explainability in the expanding field of MADRL and encourage the AAMAS community to use and engage more systematically with modern direct interpretability methods.

We list our contributions as follows:
\begin{itemize}
    \item Arguments to engage with
    direct interpretability methods.
    \item A taxonomy to position direct interpretability in MADRL.
    \item Potential applications of direct interpretability to solve modern MADRL challenges.
\end{itemize}
In this article, we first present an initial background about the systems of study and the methods advocated. Then, we propose a simple taxonomy to position modern interpretability methods in the MADRL framework. Finally, we outline the limitations of some current works while proposing alternative ideas tracks.

\section{Background}

\subsection{Multi-Agent Deep Reinforcement Learning}
\begin{figure}[ht]
    \centering
    \includegraphics[width=\linewidth]{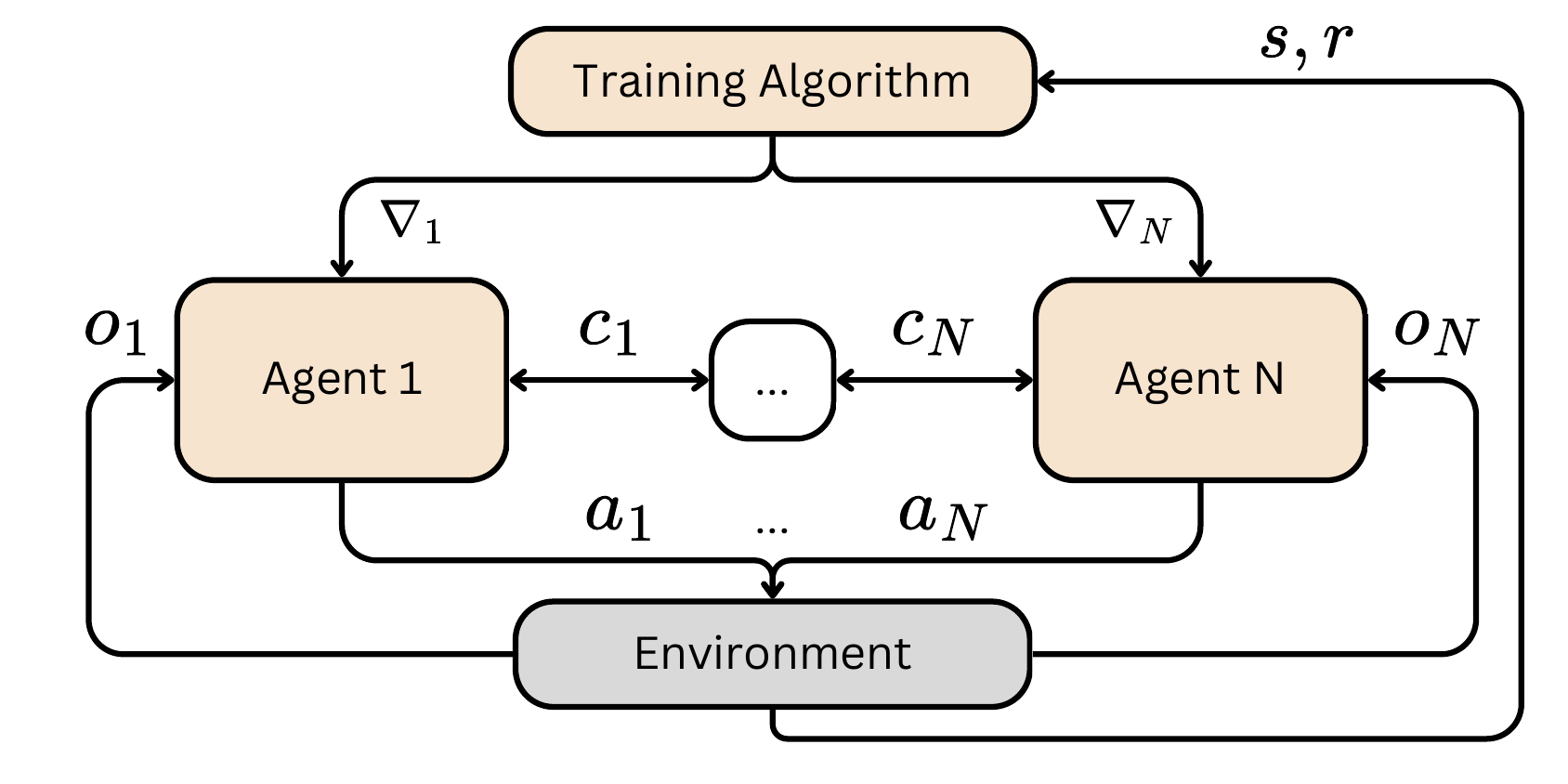}
    \caption{Schema of a simplified view of MADRL systems. At each time step, the agent $i$ receives the initial observation $o_i$, complemented by potential communications $c_i$ and produces an action $a_i$. The agent learns throughout training by the means of gradients $\nabla_i$. }
    \label{fig:madrl_system}
\end{figure}

 A typical system consists of the following components: agents, an environment, and a training algorithm, as depicted in Figure~\ref{fig:madrl_system}. Formally, we consider a system with $N$ agents, each indexed by $i \in \{1, \dots, N\}$. At each time step, the agent $i$ is presented with an observation $o_i$ and produces an action $a_i$. For the sake of generality, we included a possible communication channel $c_i$, seeing that it is increasingly used \cite{Zhu2022ASO}. In principle, we can extend the definition of communication to include the most common MADRL methods like parameter sharing \cite{Gupta2017CooperativeMC,Chu2017ParameterSD}, which can be seen as a form of latent space communication. Finally, the training algorithm provides feedback $\nabla_i$ to each agent.

Training algorithms in MADRL can be centralized, decentralized, or hybrid. Centralized training uses the joint action $a=(a_1,...,a_N)$ and the state $s$, which can be understood as an observation augmented by information at training time \cite{Lambrechts2023InformedPL}, and consists of applying classical RL to multi-agent problems like for AplhaStar \cite{Mathieu2023AlphaStarUL}. While decentralized training restricts each agent to local observations $o_i$, possibly including a local reward $r_i$, see IDQN \cite{Tampuu2015MultiagentCA} or IPPO \cite{Yu2021TheSE}. Hybrid approaches, such as centralized training with decentralized execution, leverage global information during training but allow agents to act independently using only local observations during execution, see VDN \cite{Sunehag2017ValueDecompositionNF}, QMIX \cite{Rashid2018QMIXMV}, MADPG \cite{Lowe2017MultiAgentAF} or MAPPO \cite{Yu2021TheSE}. Here, we consider agents based on DNNs; therefore, the feedbacks $\nabla_i$ are gradients of a loss $\ell$. Depending on the training algorithm, this loss can be a function of the reward $r$, the state $s$, the actions $a_i$, the observations $o_i$ and the communications $c_i$. For simplicity, we didn't include those dependencies in Figure~\ref{fig:madrl_system}.
 
\subsection{Direct Interpretability of DNNs}
\label{sec:background_interp}

We now present an overview of the modern methods widely used to interpret DNNs in Computer Vision (CV) and Natural Language Processing (NLP). As these domains heavily relied on pre-trained models \cite{Simonyan2014VeryDC,He2015DeepRL, Radford2018ImprovingLU}, direct post-hoc methods have dominated the research landscape, providing key hindsight without altering models' architectures.

\paragraph{Feature importance.} Typical methods used in CV to understand convolutional networks involve visualising important pixels, i.e. saliency maps, \cite{Zeiler2013VisualizingAU,Selvaraju2016GradCAMVE}. Other methods compute importance by perturbing the input \cite{Covert2020ExplainingBR}, using the gradients \cite{Radford2015UnsupervisedRL,Selvaraju2016GradCAMVE,Shrikumar2016NotJA, Smilkov2017SmoothGradRN} or locally decomposing relevance \cite{Montavon2015ExplainingNC,Bach2015OnPE}. Recent works in NLP focus on the Transformer architecture and its attention mechanism \cite{Vaswani2017AttentionIA}, providing token-level insights \cite{Wiegreffe2019AttentionIN,Achtibat2024AttnLRPAL}.

\paragraph{Prototypes:} a class of methods that creates explanations based on characteristic samples. In CV, it is common to analyse neurons using activation maximisation to create pre-images \cite{Mahendran2015VisualizingDC}, or find related images \cite{Chen2020ConceptWF}. Prototypes can be of various forms like perturbed images \cite{Ribeiro2018AnchorsHM}, cropped images \cite{Dreyer2023UnderstandingT} or latent space vector \cite{alain2018understanding,kim2018interpretability}. Recent works based on sparse autoencoders were able to elicit interpretable features in LLMs, i.e., prototypes \cite{Cunningham2023SparseAF}.

\paragraph{Latent manipulation:} techniques that further extend the interpretability of concepts and features by exploring the internal representations learned by models. These methods were introduced in CV with \cite{kim2018interpretability}, later derived as the field of representation engineering \cite{zou2023representation}. Such latent features enable locating, editing, erasing or decoding models' knowledge \cite{Meng2022LocatingAE,belrose2023leace, Ghandeharioun2024PatchscopesAU}, but causally modify or analyse the produced outputs \cite{rimsky2023steering, Kramar2024AtPAE}.

\paragraph{Circuit analysis:} provides a more granular understanding of model internals by examining pathways and dependencies between models' components, usually neurons or attention heads. Circuits were first discovered in CNNs \cite{Olah2020ZoomIA} before being formalised for Transformers \cite{elhage2021mathematical}.  These circuits revealed peculiar models' components that learned precise mechanisms like induction \cite{Olsson2022IncontextLA}. Using specific datasets, relevant circuits can be automatically discovered \cite{conmy2023automated}. More recent works focus on larger models' components at the layer scale \cite{Dunefsky2024TranscodersFI}.


\section{Advocating Direct Interpretability}
\label{sec:advocate}

Direct methods offer a significant advantage in their applicability to models during and after training, enabling developers to analyse and interpret complex systems without requiring architectural changes. This flexibility makes them particularly suitable for MADRL systems compared to intrinsic methods that might be challenging to scale with several agents. Figure~\ref{fig:xmadrl} outlines speculative research directions and methodologies that can enhance systems understanding at different levels, from individual agents to the overall training process.

\subsection{Single-Agent Challenges}

To understand agents trained using MADRL, we can study each agent independently.
Methods drawn from XRL and general interpretability are thus directly applicable to tackle single-agent challenges.

\paragraph{Biases identification:}eliciting models' biases learned during training. In order to debug those "Clever Hans"\footnote{Cognitive bias that was learned due to spurious correlations, see \cite{Lapuschkin2019UnmaskingCH}.}, it is possible to use feature importance techniques, described in Section~\ref{sec:background_interp}. Previous work \cite{Lapuschkin2019UnmaskingCH} showed that this debugging could be semi-automated by combining LRP \cite{Bach2015OnPE} with spectral clustering \cite{Luxburg2007ATO}.  While these methods are relatively established in XRL, further improvements tailored to MADRL could offer more context-specific explanations, e.g., by comparing different agents' perceptions.

\paragraph{Policy distillation,} converting a model into a simpler one, can be achieved by training a new smaller model \cite{Rusu2015PolicyD}, or by extracting intrinsically interpretable models \cite{Ross2010ARO,Bastani2018VerifiableRL}, even for MADRL \cite{Milani2022MAVIPERLD}. Yet, these distillation methods are computationally expensive. Recent works proposed network compression based on interpretability, using weights relevance \cite{Yeom2019PruningBE} or circuit analysis
\cite{Pochinkov2024DissectingLM}.

\paragraph{Decision decomposition:}
could be achieved by internally decomposing an agent's decision into functional modules or representations. This methodology was proven efficient to elicit the algorithms behind certain capabilities, like addition or modular addition \cite{Quirke2023UnderstandingAI,Nanda2023ProgressMF}. Future work could focus on extracting different circuits using ACDC \cite{conmy2023automated} to analyse simple shared actor-critic architectures, e.g., to extract the actor subnetwork.

\paragraph{Policy edition,}an essential aspect to regain control over DNNs. Indeed, being able to edit a trained policy is essential to remove biases, unwanted associations or dangerous behaviours without needing to retrain the model.
In this respect, direct interpretability is perfectly suited for the task with methods leveraging CAVs \cite{Dreyer2023FromHT} or causal
tracing \cite{Meng2022LocatingAE}.

\subsection{Multi-Agent Challenges}

Interpretability could be a powerful tool for automating the oversight of systems involving multiple agents.
Indeed, such systems become more complex through inter-agent interactions, coordination strategies, and emergent behaviours.

\paragraph{Team identification:} grouping together agents with similar roles or policies. This is particularly interesting to reduce the complexity of MAS by having fewer agents to train or could be an avenue to extend the mean-field framework \cite{Yang2018MeanFM}. Previous work showed that selective parameter-sharing can be based on latent spaces \cite{Christianos2021ScalingMR}. Further improvements could consider dynamic teams throughout learning by analysing mixing networks \cite{Rashid2018QMIXMV}, e.g., by partitioning the positive weights using NMF \cite{Paatero1994PositiveMF}, or using other prototype methods like SAE \cite{Cunningham2023SparseAF}.

\paragraph{Agent contribution,}or agent credit assignment, is a well-known challenge introduced by MAS. Shapley values theoretically give the individual agent contributions \cite{Shapley1988AVF}, and thus can be computed using SHAP or equivalent methods \cite{Lundberg2017AUA,Heuillet2021CollectiveEA,Wang2021SHAQIS}. Yet, as it can be expensive to compute, it might be beneficial to explore other versatile methods like LRP \cite{Bach2015OnPE}, e.g., by designing specific relevance propagation rules.

\paragraph{Communication monitoring} In settings with natural language communication between agents, leveraging LLMs or pre-trained models can enable a seamless integration \cite{Zhu2022ASO}. Yet, these models are highly opaque and would benefit from interpretability, offering an avenue to supervise and interpret conversations. Applications could make use of feature importance methods, like AttnLRP \cite{Achtibat2024AttnLRPAL}, to spot key information used in the agent prediction. 

\paragraph{Communication decoding}
For learned communication analyses, it becomes harder and might be reduced to finding patterns or comparing and aligning latent spaces to spot similar messages between agents.
In order to uncover how agents derive meaning from these interactions, causal interventions might 
yields interesting hindsights \cite{Kramar2024AtPAE}.

\paragraph{Swarm coordination:} an inherent challenge of MAS that becomes increasingly complex as the number of agents scales. Fortunately, modern direct interpretability offers means to control models using methods from representation engineering \cite{zou2023representation}, like activation steering \cite{rimsky2023steering}. The latter method has proven useful to control an agent's policy by favouring different goals \cite{Mini2023UnderstandingAC}.
Further application to MADRL could improve swarm coordination by enhancing traits like cooperativeness or better distributing goals among agents, e.g., by alternating resource collection among sites and agents.

\subsection{Training Process Challenges}
Training multiple agents simultaneously demands more computing power and can lead to learning instabilities.
Therefore, it is crucial to better understand the training process of MADRL at different levels by improving learning efficiency and ensuring robustness. 

\paragraph{State analysis.}In order to model complex environments, one can train world models \cite{Bruce2024GenieGI}, later used by an agent \cite{Hafner2023MasteringDD}. 
The condensed latent representation obtained can be analysed \cite{Ivanitskiy2023StructuredWR}
with tools like the tuned lens \cite{Belrose2023ElicitingLP}. This framework offers a better view of the transition function, which could help guide the agents towards unbiased training if analysed thoroughly.

\paragraph{Reward decomposition:} often achieved by learning separate value functions aggregated afterwards \cite{Seijen2017HybridRA,Juozapaitis2019ExplainableRL}. 
To avoid arbitrary decompositions, one could rely on local backpropagation methods like LRP or CRP \cite{Bach2015OnPE,Achtibat2022FromAM}, enabling the discovery of concepts that can later clarify the influence of the reward on the learning process of a policy.
Further improvements could consider generating an adaptative curriculum \cite{Jiang2020PrioritizedLR}, prioritizing the concepts to learn.

\paragraph{Priority sampling,} a staple method in RL that improves sample efficiency \cite{Schaul2015PrioritizedER}. 
Also, in RL, interpretability was proven efficient to prioritize the important pixels for a visual policy by means of a consistency loss \cite{Bertoin2022LookWY}. Such a framework could be extended to compute importance over multiple inputs, creating a metric for better eliciting shared critical training samples.  

\paragraph{Learning dynamics:} trying to understand the agents throughout training, e.g., by observing the trained policies. Yet, it becomes more complicated as the number of agents scales and requires automated methods beyond observing policies. A widely used method to detect learned concepts in a model is to train linear probes
\cite{alain2018understanding}, which gave valuable insights for the analysis of AlphaZero networks \cite{McGrath_2022}.
By monitoring each agent, it would be possible to gain a more nuanced understanding of the swarm development and track the emergence or disappearance of certain capabilities.

\paragraph{Experience sharing:} a method introduced to scale MADRL by improving sample efficiency \cite{Christianos2020SharedEA}. Further improvements shared the data selectively according to exploration metrics \cite{Gerstgrasser2023SelectivelySE}. Yet, this framework is missing a key point: you might want to select agents that share their experience similarly to parameter sharing \cite{Christianos2021ScalingMR}. A naive method could be to cluster experiences based on some latent representation of the different agents, enabling efficient knowledge sharing \cite{zou2023representation}.

\section{Discussion}

\subsection{Post-Hoc Interpretability in Deep RL}

Post hoc interpretability in Deep Reinforcement Learning (DRL) is an increasingly important field, with methods such as saliency maps already being used to visualize agent behaviour \cite{Greydanus2017VisualizingAU}, debug learned concepts \cite{Jaderberg2018HumanlevelPI, Hilton2020UnderstandingRV}, and inform sampling strategies to improve efficiency \cite{Bertoin2022LookWY}. Other approaches analyse agent behaviour by querying interaction data \cite{Sequeira2019InterestingnessEF} or by visualising pattern prototypes \cite{Ragodos2022ProtoXEA,Aliciolu2024UseBA}.  More extensive efforts have focused on interpreting well-known chess engines like AlphaZero \cite{McGrath_2022,lovering2022evaluation,hammersborg2023information,schut2023bridging,jenner2024evidencelearnedlookaheadchessplaying,Poupart2024ContrastiveSA} and Stockfish \cite{palsson2023unveiling}, providing valuable insights into learned strategies. Ongoing efforts are also focused on exposing the key mechanisms behind planning, especially with games as a testbed \cite{Jenner2024EvidenceOL,Taufeeque2024PlanningIA,Guei2024InterpretingTL,Chung2024PredictingFA}.

Other post hoc methods, like policy distillation into interpretable models, often referred to as model extraction, have also been a central focus. Techniques such as DAGGER \cite{Ross2010ARO} and VIPER \cite{Bastani2018VerifiableRL} leverage imitation learning to simplify policies. However, these methods struggle to scale effectively when applied globally to complex models, limiting their applicability to large-scale systems.

\subsection{Interpretability in MADRL}

Interpretability in MADRL is an evolving field with several promising approaches. Shapley values have been widely applied to analyse individual agent contributions, providing a robust theoretical framework for evaluating each agent’s influence on team performance \cite{Heuillet2021CollectiveEA,Wang2021SHAQIS,Mahjoub2023EfficientlyQI}. Diversity measures of agent policies have also emerged as a valuable tool for understanding agent behaviour, revealing distinctions between individual strategies and their roles in collective dynamics \cite{Khlifi2023OnDF}.

Similarly to XRL, policy extraction techniques, such as VIPER \cite{Bastani2018VerifiableRL}, have been extended to leverage MADRL training to distil interpretable policies from complex models \cite{Milani2022MAVIPERLD}. Furthermore, predicting high-level concepts instead of actions offers a novel pathway to intrinsically interpretable models, aligning model outputs with human-understandable abstractions \cite{Zabounidis2023ConceptLF}. These advancements highlight the growing potential of interpretability methods in uncovering insights into multi-agent behaviour and learning processes.

\subsection{Limits of Intrinsically Interpretable Models}

Intrinsically interpretable models, whether obtained by design or post hoc extraction, have long been a dominant paradigm in agent interpretability research, relying on predefined, transparent model architectures. Design frameworks like XAg \cite{rodriguez2024explainable}, concept bottlenecks \cite{Poeta2023ConceptbasedEA}, learning skills with decision trees \cite{Wen2024SkillTreeES}, or learning modularised agents \cite{Cloud2024GradientRM}, aim to embed interpretability directly into model structures. However, such approaches face challenges in scalability and flexibility, particularly in multi-agent settings or with complex DRL models like the latest pre-trained world models \cite{Reed2022AGA,Yang2023FoundationMF, alonso2024diffusionworldmodelingvisual, Bruce2024GenieGI}. The rigidity of design-based interpretability often compromises performance and fails to capture emergent behaviours, highlighting the need for alternative approaches that can adapt to the complexity and scale of modern systems \cite{Madsen2024InterpretabilityNA}. New hybrid paradigms like Wrapper Boxes \cite{Su2023InterpretableBD}, might be required to overcome those limitations.

\section{Perspectives}

\subsection{MADRL Should Leverage Direct Interpretability}

Engaging and expanding interpretability is an opportunity to address existing challenges in MADRL. Direct approaches are particularly well-suited for analysing communication dynamics, coordination strategies, and emergent behaviours in MAS. Graph-based analysis, for instance, could provide insights into inter-agent interactions, while feature importance techniques can identify biases and ensure fairness in decision-making. By systematically exploring and applying scalable direct methods to trained models, researchers can better address the inherent complexities of MADRL, enabling the development of more transparent, robust, and accountable systems for real-world applications.

Although previous calls to action are prone to integrate interpretability beforehand \cite{rodriguez2024explainable}, this paper claims that the interpretation of models post hoc is highly valuable. Direct interpretability offers greater flexibility, particularly for existing models where architectural modifications are impractical. 

\subsection{Robust Evaluation Protocols}

As repeatedly outlined, direct post-hoc methods are easily actionable and scalable.
However, their adoption requires acknowledging and addressing limitations such as the inherent shortcomings of saliency maps \cite{Adebayo2018SanityCF,Bilodeau2022ImpossibilityTF}, counterfactual explanations \cite{Laugel2019TheDO}, or other interpretability illusions \cite{Bolukbasi2021AnII,Friedman2023InterpretabilityII,Friedman2023InterpretabilityII}. In fact, these methods often generate metrics with limited predictive power, and thus, claims should be reasonable.

A key priority is the development of robust evaluation protocols for direct methods. Given the absence of ground-truth explanations, reliable metrics and standardized evaluation frameworks must be established to assess the quality and utility of these methods \cite{Gill2020ARM,Madsen2021PosthocIF,Amorim2023EvaluatingPI,Hedstrm2022QuantusAE,Wei2024RevisitingTR,Huang2024RAVELEI,Chaudhary2024EvaluatingOS}. 
Advancing evaluation thoroughly, e.g., by evaluating out of distribution, is especially important to develop scalable, effective, and actionable interpretability solutions.

\section{Conclusion}

We outlined that direct interpretability might be vital for addressing the challenges of scalability and complexity in modern MADRL. It enables the analysis of trained models without imposing architectural constraints, providing critical insights into agent behaviour, emergent dynamics, and biases. Advancing these methods will ensure scalable oversight of these systems, which is a precious desideratum for real-world applications. However, challenges such as explanation illusions, lack of robust evaluation metrics, and difficulty disentangling causal effects should be considered and tackled.



\bibliographystyle{ACM-Reference-Format} 
\bibliography{main}


\end{document}